\documentclass[sigconf]{acmart}
\usepackage{multirow}
\usepackage[ruled,vlined]{algorithm2e}
\usepackage{tabularx}

\AtBeginDocument{%
  }

\setcopyright{acmlicensed}
\copyrightyear{2024}
\acmYear{2024}
\acmDOI{XXXXXXX.XXXXXXX}

\acmConference[Arxiv]{}{2024}{Montreal, Canada}
\acmISBN{978-1-4503-XXXX-X/18/06}

\newcommand{\B}{\textbf}

\begin{document}

\title{E2ESlack: An End-to-End Graph-Based Framework for Pre-Routing Slack Prediction}

\author{Saurabh Bodhe}
\affiliation{
  \institution{Huawei Noah's Ark Lab}
  \city{Montreal}
  \country{Canada}
}
\email{}

\author{Zhanguang Zhang}
\affiliation{
  \institution{Huawei Noah's Ark Lab}
  \city{Montreal}
  \country{Canada}
}
\email{}

\author{Atia Hamidizadeh}
\affiliation{
  \institution{Huawei Noah's Ark Lab}
  \city{Montreal}
  \country{Canada}
}
\email{}

\author{Shixiong Kai}
\affiliation{
  \institution{Huawei Noah's Ark Lab}
  \city{Beijing}
  \country{China}
}
\email{}

\author{Yingxue Zhang}
\affiliation{
  \institution{Huawei Noah's Ark Lab}
  \city{Montreal}
  \country{Canada}
}
\email{}

\author{Mingxuan Yuan}
\affiliation{
  \institution{Huawei Noah's Ark Lab}
  \city{Hong-Kong}
  \country{Hong Kong}
}
\email{}

\renewcommand{\shortauthors}{Bodhe, Zhang et al.}

\begin{abstract}
  Pre-routing slack prediction remains a critical area of research in Electronic Design Automation (EDA). Despite numerous machine learning-based approaches targeting this task, there is still a lack of a truly end-to-end framework that engineers can use to obtain TNS/WNS metrics from raw circuit data at the placement stage. Existing works have demonstrated effectiveness in Arrival Time (AT) prediction but lack a mechanism for Required Arrival Time (RAT) prediction, which is essential for slack prediction and obtaining TNS/WNS metrics. In this work, we propose E2ESlack, an end-to-end graph-based framework for pre-routing slack prediction. The framework includes a TimingParser that supports DEF, SDF and LIB files for feature extraction and graph construction, an arrival time prediction model and a fast RAT estimation module. To the best of our knowledge, this is the first work capable of predicting path-level slacks at the pre-routing stage. 
  We perform extensive experiments and demonstrate that our proposed RAT estimation method outperforms the SOTA ML-based prediction method and also pre-routing STA tool. Additionally, the proposed E2ESlack framework achieves TNS/WNS values comparable to post-routing STA results while saving up to 23x runtime. 

\end{abstract}

\ccsdesc[500]{Hardware~Static timing analysis}
\ccsdesc[300]{Hardware~Wire routing}
\ccsdesc[500]{Hardware~Software tools for EDA}

\keywords{timing, gnn, arrival time, TNS/WNS, STA, TimingParser, OpenLane}

\received{12 July 2024}

\maketitle

\section{Introduction}

In the circuit design process, Total Negative Slack (TNS) and Worst Negative Slack (WNS) are two critical metrics used to evaluate circuit quality. Slacks are obtained through Static Timing Analysis (STA), which is performed multiple times during the Electronic Design Automation (EDA) process. Accurate slack can only be calculated after routing when the parasitics of interconnects are available. However, routing is often the most expensive step of design flow and the cost of design iteration after routing is high. With the evolution of "shift-left" in EDA industry \cite{bhardwaj2021shift}, it is desirable to evaluate slacks in circuits at early design stages to enable faster design iteration and reduce Time-to-Market (TTM) cost.


Analytical placement tools like ePlace \cite{lu2015eplace} and DREAMPlace \cite{lin2019dreamplace} use half-perimeter wirelength (HPWL) as the surrogate of design timing quality for placement optimization. With the advancement of machine learning (ML) techniques, multiple data-driven methods have been proposed to predict timing metrics at placement stage \cite{barbozapreroute, guo2022timing, zhong2024preroutgnn} or logic synthesis stage \cite{sengupta2022iccad, fang2023masterrtl}. Computation of slack requires both Arrival Time (AT) and Required Arrival Time (RAT) at the endpoint of each timing path. Prior timing prediction methods, such as TimingPredict \cite{guo2022timing} and PreRoutGNN \cite{zhong2024preroutgnn}, are capable of predicting AT at the pre-routing stage, but they lack a method of acquiring pre-routing RAT. PreRoutGNN \cite{zhong2024preroutgnn} assumes that pre-routing RAT can be obtained from the Synopsys Design Constraints (SDC) file, which is not entirely valid since the SDC file only contains RAT for primary
outputs, but not for other endpoints. Due to backward propagation mechanism of RAT calculation, their AT prediction model cannot be directly used for RAT prediction. Without RATs of endpoint pins, pre-routing slacks cannot be obtained, limiting the potential use of their method as an early indicator of circuit timing performance. To overcome this challenge, we present a fast RAT estimation algorithm that leverages AT predictions from existing models to provide an accurate estimate of RAT. This contributes to a comprehensive framework for accurate prediction of slack and TNS/WNS at the pre-routing stage.

Our contributions are summarized as follows:
\begin{itemize}
\item \textbf{E2ESlack} - We propose an end-to-end slack prediction framework at the pre-routing stage and validate its performance through extensive experiments. The framework achieves TNS/WNS values comparable to post routing STA results while reducing runtime by up to 23 times.
\item \textbf{TimingParser} - We introduce a fast, distributed framework for converting raw circuit data (SDF, LIB, DEF) to PyTorch DGL graphs. We utilize the Ray multiprocessing library~\cite{moritz2018ray} for acceleration.
\item \textbf{RAT Estimation} - We develop a novel algorithm for fast estimation of RAT, which enables prediction of TNS and WNS at the pre-routing stage. To the best of our knowledge, this is the first work addressing the pre-routing RAT estimation task. 
\end{itemize}

\begin{figure*}[t]
  \centering
  \includegraphics[width=0.95\textwidth]{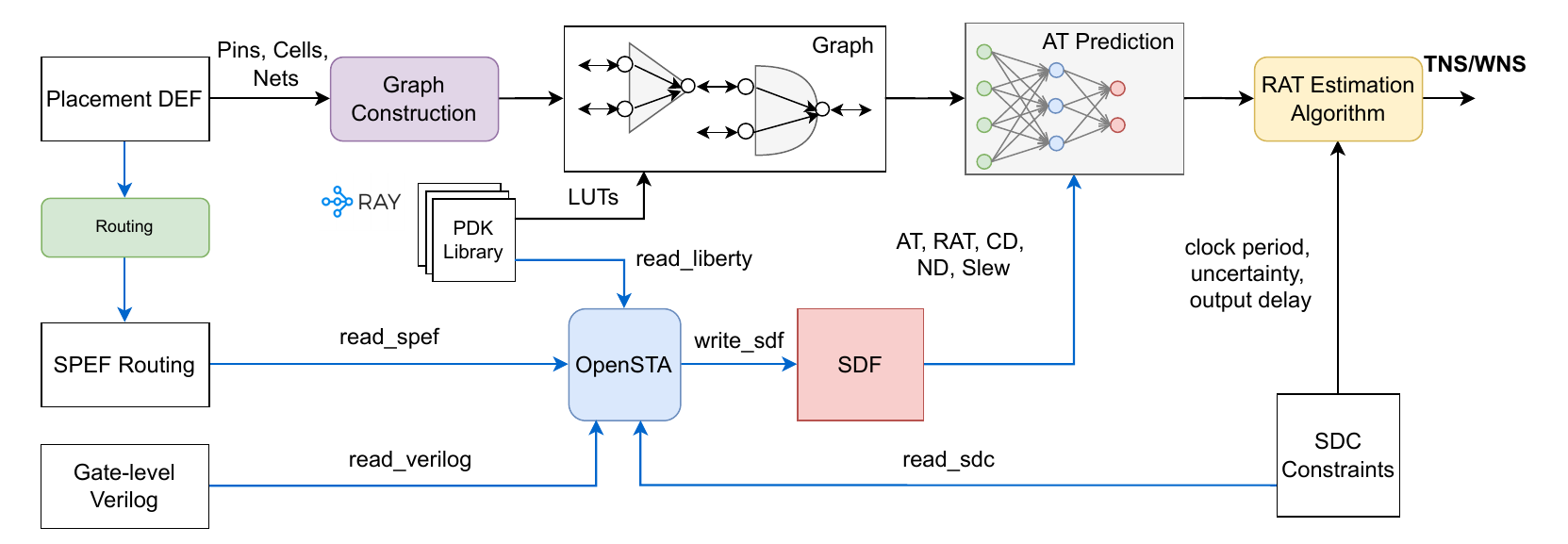}
  \caption{Framework Architecture. Blue arrows indicate the workflow for label generation, which is only required for training. Black arrows indicate the inference workflow.}
  \Description{}
  \label{fig:arch}
\end{figure*}

\section{Related Work \& Preliminaries}

\subsection{Static Timing Analysis}
In circuit design workflow, Static Timing Analysis (STA) is performed at various stages to evaluate the timing performance and check setup or hold violations \cite{mcwilliams1980verification}. Post-routing STA should be done under four corner conditions: early-late/rise-fall (EL/RF) \cite{friedman2001clock}. STA tools utilize the slack value to pinpoint timing path violations for further optimization. 
Slack is defined as the difference between the Required Arrival Time (RAT) and Arrival Time (AT) of the data signal at the endpoint of a timing path:
$$
Slack_L = RAT_L - AT_L, Slack_E = AT_E - RAT_E
$$
where $L$ means late or setup slack and $E$ means early or hold slack. A negative slack indicates violation of setup or hold constraints. While both setup and hold slacks are important and need to be checked in STA tools, we focus on estimating slack at the late corner or setup slack which is more useful for timing-driven placement optimization.



\subsection{Pre-Routing Slack Prediction}
 Machine learning algorithms have been employed in various circuit design tasks~\cite{chan2016learning, guo2022timing, han2014deep, liu2021parasitic, li2020customized, liang2020routing}. Given that circuits can naturally be represented as graphs, graph neural networks (GNNs) have gained popularity in EDA research~\cite{liu2021parasitic, xie2021net2, yang2022versatile, guo2022timing}. However, typical GNN models are subject to oversmoothing issue and will perform poorly on slack prediction task due to the high topological level of circuits (around 300 on large designs with millions pins).  
 To address this issue, TimingPredict~\cite{guo2022timing} proposes a custom GNN model inspired by timing engines for Arrival Time (AT) prediction. A subsequent work PreRoutGNN~\cite{zhong2024preroutgnn} proposes a pre-training approach along with topological level encoding to reduce the error accumulation issue of TimingPredict. Although these two methods can predict AT precisely, they lack a method to acquire Required Arrival Time (RAT) at the pre-routing stage. TimingPredict~\cite{guo2022timing} uses post-routing RATs for slack calculation, which is not ideal since the goal is to make pre-routing predictions without running expensive routing tools. PreRoutGNN~\cite{zhong2024preroutgnn} assumes that RAT is available in external SDC files, which is not an entirely valid assumption as the SDC file only contains RAT for primary outputs but not for timing paths the end with registers. In this work, we propose an algorithm to estimate RAT as part of the pre-routing slack prediction framework. 

\section{Framework}

\subsection{Graph Representation}
To the best of our knowledge, TimingPredict~\cite{guo2022timing} remains the SOTA model for AT prediction using circuit graphs that is also open source. Hence, we choose to use a representation identical to the one proposed in that work. This enables the use of their model as-is in our pipeline. However, the modular structure of our work enables easy adaptation to future AT prediction models that use graph-based methods.

The circuit is represented as a directed heterogeneous graph, where each pin of the circuit is treated as a graph node. The two types of timing arc are represented as edges, namely net edges and cell edges. The net edges are bidirectional, while the cell edges are unidirectional. Each net edge represents the connection between a source and sink of a net. The cell edges are inside the cell and represent the timing arc directed from input pin of a given cell to its output pin. Our implementation also includes cycle detection and removal. Table \ref{tab:features} summarizes the features for the nodes and edges.

\begin{table}
  \resizebox{.45\textwidth}{!}{%
  \begin{tabular}{|c|c|c|}
    \toprule
    Type&Attribute&Description\\
    \midrule
    \multirow{6}{4em}{Pin node} & is primary I/O & boolean indicating if pin is primary I/O\\
    &location of pin&distance of pin from chip boundaries\\
    &pin capacitance&EL/RF capacitance obtained from cell library\\
    &direction&if the pin is fan-in or fan-out \\
    &arrival time&AT label for model training only \\
    &slew&Slew label for model training only \\
    &&\\
    \multirow{3}{4em}{Net edge}&\multirow{2}{3em}{length}&Manhattan distance between \\
    &&2 pins connected by net\\
    &net delay&ND label for model training only \\
    &&\\
    \multirow{4}{4em}{Cell edge}&is valid LUT&boolean to indicate if the LUT is valid\\
    &LUT indices&EL/RF LUT indices from cell library \\
    &LUT values&EL/RF LUT values from cell library \\
    &cell delay&CD label for model training only\\
  \bottomrule
\end{tabular}}
\vspace{1em}
\caption{Node and edge features}
\label{tab:features}
\end{table}

\begin{table}[t]
  \resizebox{.45\textwidth}{!}{%
  \begin{tabular}{|c|c|c|c|c|}
    \toprule
    Benchmark & \#Nodes&\# Cell edges& \# Net Edges&\# Endpoints\\
    \midrule
    aes& 82371 & 54159 & 57874 & 3026\\
    aes$\_$core & 97111& 50427& 73441& 2606\\
    APU&12083&7984	&8447	&444\\
    gcd	&1306	&835	&894	&53\\
    inverter	&8	&3	&4	&1\\
    PPU	&48437	&29082	&32402	&2972\\
    s44	&523	&294	&335	&41\\
    chacha	&58582	&38393	&41398	&1972\\
    ldpcenc	&98264&	69315	&73736	&1400\\
    md5&	23860& 	15361&	16611&	1041\\
    ocs$\_$blitter&	33440	&23516	&24423	&552\\
    point$\_$mult&	187498&	116739&	134186&	6509\\
    y$\_$dct	&294438&	238089&	183429&	6044\\
  \bottomrule
\end{tabular}}
\vspace{1em}
\caption{Statistics for the proposed OpenLane benchmark}
\label{tab:stats}
\end{table}

\subsection{TimingParser}

As part of our work, we propose a fast, distributed, and modularized parser. It can convert a circuit represented by LIB, SDF, and DEF files to a PyTorch DGL graph. We accelerate it using Ray library~\cite{moritz2018ray} for multiprocessing. It also includes support for handling circuits that use multiple PDK libraries and for variable-sized LUTs. These features are essential for industry-grade designs.

Liberty, also called LIB is a standard format developed by Synopsis for representing timing and power characteristics of standard cells. For our work, we are interested in extracting the pin capacitance and timing lookup tables (LUT) that are used as input features as shown in Table \ref{tab:features}.

All circuits in the proposed OpenLane dataset as well as existing TimingPredict~\cite{guo2022timing} dataset use the SkyWater 130nm process design kit (PDK). Hence we use the pair of early and late libraries \texttt{sky130\_fd\_sc\_hd\_\_ff\_n40C\_1v95.lib} and \texttt{sky130\_fd\_sc\_hd\_\_ ss\_100C\_1v60.lib} respectively. We extract the 4 tables named \texttt{cell\_fall}, \texttt{cell\_rise}, \texttt{fall\_transition} and \texttt{rise\_transition}. Each table consists of two indices for the two dimensions of the table and the corresponding value matrix. From both libraries together, we get a set of 8 tables in total for each cell class. These are assigned as cell edge features for all cells belonging to that particular class. In case of SkyWater libraries all LUTs are of the same size 7 x 7, but this is not necessarily true for all libraries. A design might also use multiple libraries with LUTs of different sizes, but the prediction model cannot accept variable-sized LUTs as input features. To handle this we implement a mechanism to linearly interpolate smaller LUTs to a fixed configurable size. This is inline with actual mechanism of timing engines since STA tools do lookup operation for intermediate indices by interpolating. We implement multiprocessing for parsing multiple libraries using the Ray framework~\cite{moritz2018ray}.

Design Exchange Format (DEF) is a standard format developed by Cadence to represent the physical layout of a circuit in ASCII format. We obtain the location of 4 corners of the die from the 'DIEAREA' line. From the 'PINS' section we obtain detailed information about the primary inputs (PI) and primary outputs (PO), including their location as coordinates on the die. In the 'COMPONENTS' section we can find a list of all cells in the design including the cell name, cell class corresponding to the PDK library and the cell location. Finally the 'NETS' section contains information about how the cells are connected. Each net lists all the pins which are connected together. Exactly one of the pins in the list is net source while others are net sinks.

Standard Delay Format (SDF) is an IEEE standard for representing timing data at any stage in EDA flow. We use this only to obtain labels for AT, RAT, Cell Delay (CD), Net Delay (ND), and Slew. These labels are used to train the AT prediction model. We use the \texttt{'write\_sdf'} command of OpenSTA~\cite{opensta} tool to obtain the SDF file. By default, this SDF contains only net delays and cell delays. Hence, we use a modified version of OpenSTA~\cite{openstadump} developed by authors of TimingPredict~\cite{guo2022timing} which is also able to include AT, RAT, and Slew for all pins in the SDF.

\subsection{TNS/WNS Estimation Algorithm}

\begin{algorithm}
\SetAlgoLined
\SetKwInOut{Input}{Input}
\SetKwInOut{Output}{Output}
\Input{Graph, Predicted AT, SDC Constraints}
\Output{RAT for each endpoint, TNS/WNS for overall circuit }
endpoints $\leftarrow$ nodes\_with\_zero\_out\_degree()\;
 \For{each endpoint}{
  \If{endpoint is primary output}{
   RAT(endpoint) = clk\_period – output\_delay(endpoint) – clk\_uncertainty\;
  }
  \Else{
   RAT(endpoint) = clk\_period + min\_clk\_delay(endpoint) – clk\_uncertainty\;
  }
 }
 critical\_paths $\leftarrow$ [ ]\;
 \For{each endpoint}{
  Slack(endpoint) = RAT(endpoint) – AT$_{pred}$(endpoint)\;
  \If{Slack(endpoint) $<$ 0}{
   critical\_path = \textbf{BackwardTraversalTillCLK}(endpoint)\;
   critical\_paths.append(critical\_path)\;
  }
 }
\For{each (critical\_path, endpoint)}{
  startpoint = \textbf{ForwardTraversalTillRegister}(critical\_path)\;
  sp\_clock\_path = get\_clock\_path(startpoint)\;
  ep\_clock\_path = get\_clock\_path(endpoint)\;
  common\_path = get\_common\_path(sp\_clock\_path, ep\_clock\_path)\;
  CRP(endpoint) = max\_delay(common\_path) – min\_delay(common\_path)\;
  RAT$_{corrected}$(endpoint) = RAT(endpoint) + CRP(endpoint)\;
  Slack$_{corrected}$(endpoint) = RAT$_{corrected}$(endpoint)  –  AT$_{pred}$(endpoint)\;
 }
 TNS $\leftarrow$ sum(negative$\_$slack)\;
 WNS $\leftarrow$ min(negative$\_$slack)\;

 generate$\_$report()\;
 \caption{Algorithm for TNS and WNS calculation}
\end{algorithm}

\begin{table*}[t]
  \centering
  \resizebox{.95\textwidth}{!}{%
  \begin{tabular}{|c|c|c|ccc|cccc|ccc|c|}
  \toprule
    & & \multicolumn{2}{c}{TimingPredict\cite{guo2022timing}
    } & PreR STA & Ours & \multicolumn{7}{c|}{Runtime (sec)} & \\
    & Circuit & AT R2 & RAT MAE & RAT MAE & RAT MAE & Parser & AT Pred & RAT & Total & Routing & STA & Total & Speedup\\
    \midrule
    \multirow{23}{2em}{train} & blabla & 0.9437 & 27.8952 & 1.9292 & \B{0.0966} & 26.98 & 0.57 & 0.09 & \B{27.64} & 836 & 23.6 & 859.6 & 31$\times$\\
    & usb$\_$cdc$\_$core & 0.9785 & 2.4454 & \B{0.2291} & 0.2720 & 16.33 & 0.11 & 0.01 & \B{16.45} & 3654 & 4.7 & 3658.7 &222$\times$ \\
    & BM64 & 0.9643 & 6.1677 & 1.0263 & \B{0.3018} & 23.02 & 0.16 & 11.1 & \B{34.28} & 713 & 13.6 & 726.6 & 21 $\times$\\
    & salsa20 & 0.9702 & 11.7409 & 2.337 & \B{0.2305} & 31.03 & 0.27 & 1.68 & \B{33.98} & 1738 & 29.1 & 1767.1 & 52$\times$\\
    & aes128 & 0.9316 & 3.3494 & 3.4672 & \B{1.9977} & 52.66 & 0.28 & 22.02 & \B{74.96} & 1787 & 51.4 & 1838.4 &25$\times$\\
    & wbqspiflash & 0.9842 & 3.6588 & 0.3689 & \B{0.1364} & 17.39 & 0.17 & 0.08 & \B{17.64} & 181 & 5.9 & 186.9 & 11$\times$\\
    & cic$\_$decimator & 0.9689 & 0.6819 & 0.2188 & \B{0.1171} & 16.27 & 0.06 & 0.01 & \B{16.34} & 81 & 3.3 & 84.3 & 5$\times$\\
    & aes256 & 0.948 & 4.6155 & 4.6156 & \B{3.6112} & 66.21 & 0.28 & 31.47 & \B{97.96} & 2886 & 72.0 & 2958 & 30$\times$\\
    & des & 0.9702 & 4.2351 & \B{0.7797} & 2.5621 & 26.61 & 0.09 & 7.4 & \B{34.1} & 493 & 16.2 & 509.2 & 15$\times$\\
    & aes$\_$cipher & 0.9554 & 3.5156 & \B{0.6617} & 0.6713 & 25.71 & 0.12 & 2.57 & \B{28.40} & 1849 & 18.7 & 1867.7 & 66$\times$\\
    & picorv32a & 0.8578 & 8.3143 & 0.9441 & \B{0.6909} & 26.33 & 0.27 & 1.03 &\B{ 27.63} & 987 & 18.5 & 1005.5 &36$\times$\\
    & zipdiv & 0.9791 & 2.3757 & 0.3931 & \B{0.1621} & 16.41 & 0.11 & 0.01 & \B{16.53} & 48 & 3.5 & 51.5 &3$\times$\\
    & genericfir & 0.6371 & 2.252 & 1.3937 & \B{0.1376} & 25.42 & 0.06 & 0.17 & \B{25.65} & 409 & 11.1 & 420.1 &16$\times$\\
    & usb & 0.9663 & 1.0913 & 0.2060 & \B{0.1625} & 15.93 & 0.05 & 0.01 & \B{15.99} & 45 & 3.4 & 48.4 &3$\times$\\
    \cmidrule{2-2}
    & aes & 0.9693 & 7.5059 & 0.9704 & \B{0.7174} & 37.92 & 0.12 & 0.11 & \B{38.15} & 442 & 18.8 & 460.8 & 12$\times$\\
    & aes$\_$core & 0.9742 & 16.3698 & 0.7592 & \B{0.1401} & 36.93 & 0.13 & 0.21 & \B{37.27} & 532 & 24.1 & 556.1 & 15$\times$\\
    & APU & 0.9845 & 1.7673 & 0.2987 & \B{0.1896} & 18.33 & 0.09 & 0.07 & \B{18.49} & 82 & 4.0 & 86 &5$\times$\\
    & gcd & 0.8495 & 4.9599 & 0.2780 & \B{0.0962} & 16.42 & 0.04 & 0.01 & \B{16.47} & 16 & 1.1 & 17.1 &1$\times$\\
    & inverter & 0.9798 & 0.121 & 0.0338 & \B{0.0} & 16.48 & 0.02 & 0.02 & 16.52 & 1 & 0.9 & \B{1.9} & 0$\times$\\
    & PPU & 0.9827 & 5.6569 & 1.3257 & \B{0.1558} & 27.66 & 0.10 & 0.11 & \B{27.87} & 161 & 13.2 & 174.2 & 6$\times$\\
    & s44 & 0.9977 & 0.6684 & 0.1843 & \B{0.1114} & 15.06 & 0.03 & 0.01 & 15.10 & 5 & 1.0 & \B{6} & 0$\times$\\
    & chacha & 0.9906 & 8.6985 & 1.1749 & \B{0.2786} & 28.68 & 0.19 & 2.6 & \B{31.47} & 377 & 14.6 & 391.6 &12$\times$\\
    & ldpcenc & 0.7913 & 5.4378 & \B{0.7686} & 0.8026 & 43.15 & 0.10 & 0.21 & \B{43.46} & 532 & 21.1 & 553.1 & 13$\times$\\
    \cmidrule{2-14}
    & \B{Average} &0.9380&	5.8054& 1.0593& \B{0.5931}&	27.2578	&0.15	&3.52&	\B{30.93}&	776&	16.3&	792.6 &26$\times$\\
    \midrule
    \multirow{12}{2em}{test} & jpeg$\_$encoder & 0.9169 & 7.0445 & 2.9086 & \B{1.7042} & 60.75 & 0.27 & 63.59 & \B{124.61} & 4191 & 70.3 & 4261.3 & 34$\times$\\
    & usbf$\_$device & 0.9202 & 5.039 & 1.4754 & \B{0.2815} & 27.38 & 0.19 & 0.99 & \B{28.56} & 1525 & 21.0 & 1546 &54$\times$\\
    & aes192 & 0.953 & 3.3412 & 4.5551 & \B{3.1170} & 57.76 & 0.28 & 26.23 & \B{84.27} & 1728 & 58.6 & 1786.6 &21$\times$\\
    & xtea & 0.9534 & 9.9293 & 0.8966 & \B{0.1566} & 17.00 & 0.18 & 0.07 & \B{17.25} & 174 & 4.9 & 178.9 &10$\times$\\
    & spm & 0.9605 & 6.5336 & \B{0.0450} & 0.1409 & 14.80 & 0.02 & 0.01 & \B{14.83} & 14 & 4.2 & 18.2 & 1$\times$\\
    & y$\_$huff & 0.8231 & 2.7866 & 1.4530 & \B{0.1504} & 24.44 & 0.10 & 0.3 & \B{24.84} & 1162 & 15.4 & 1177.4 & 47$\times$\\
    & synth$\_$ram & 0.8911 & 23.6802 & 0.9941 & \B{0.1299} & 19.85 & 0.04 & 0.04 & \B{19.93} & 452 & 9.3 & 461.3 & 23$\times$\\
    \cmidrule{2-2}
    & md5 & 0.9059 & 11.8095 & 0.8160 & \B{0.1487} & 20.58 & 0.17 & 0.05 & \B{20.8} & 141 & 5.8 & 146.8 & 7$\times$\\
    & ocs$\_$blitter & 0.9474 & 6.0162 & 0.5449 & \B{0.4137} & 23.55 & 0.14 & 0.06 & \B{23.75} & 342 & 9.6 & 351.6 & 15$\times$\\
    & point$\_$mult & 0.7279 & 1.7716 & 1.3465 & \B{0.4487} & 54.50 & 0.13 & 3.89 & \B{58.52} & 1066 & 41.4 & 1107.4 & 19$\times$\\
    & y$\_$dct & 0.7854 & 11.8169 & 1.6458 & \B{0.4080} & 93.21 & 0.25 & 8.61 & \B{102.07} & 975 & 59.6 & 1034.6 &10$\times$\\
    \cmidrule{2-14}
    & \B{Average}& 0.8895&	8.1608&	1.5165& \B{0.6454}&	37.62&	0.16&	9.44&	\B{47.22}&	1070&	27.3& 1097.3 & 23$\times$\\
    \bottomrule
  \end{tabular}}
  \vspace{1em}
  \caption{Comparison of RAT estimation performance (MAE) for circuit endpoints and runtime (sec). Our RAT algorithm is compared with modified TimingPredict~\cite{guo2022timing} and pre-routing STA. Post-routing STA is used as the ground truth in all cases. Runtime of our framework consisting of parsing, AT prediction and RAT estimation is compared to traditional method of Routing and STA using OpenLane~\cite{openlane}}
  \label{tab:results}
\end{table*}

Total Negative Slack (TNS) and Worst Negative Slack (WNS) are important metrics that can help engineers evaluate a circuit for timing violations after static timing analysis (STA). We can define a critical path as a timing path which has negative slack at its endpoint. Then TNS is simply the sum of the slacks of the endpoints of all the critical paths. WNS is the worst or the minimum slack across the endpoints of critical paths. Both these quantities are negative or zero. Intuitively, TNS tells us how bad the circuit is overall, while WNS tells us how bad the worst path of the circuit. TNS and WNS equal to zero indicate that there are no timing violations and the circuit design process can move on to the next steps.
Prior work\cite{guo2022timing, zhong2024preroutgnn} does not have a mechanism for either estimating or predicting RAT which would force engineers to rely on inaccurate pre-routing STA. We propose an algorithm which enables fast estimation of RAT given the circuit graph and the predicted ATs at each pin. RAT is defined at primary outputs as follows:
$$
RAT^{PO} = T - D^{PO} - \mu
$$
Whereas for other endpoints at synchronous elements (denoted by S) like flip flops and latches, we can define RAT as:
$$
RAT^{S} = T + D^{S}_{E}\\ - \mu + \text{CRP}
$$
Here $T$ = clock period, $\mu$ = clock uncertainty, $D^{PO}$ = output delay at PO. These can be obtained from the external design constraints SDC file. 

$D^{S}_{E}$ is the minimum (early) delay of the clock path for synchronous element S. To estimate this we can use the  predicted AT at CLK pin of element S. CRP stands for Clock Reconvergence Pessimism. RAT calculation also involves another factor not shown in formulae above, cell setup time ($T_{su}$) which is typically obtained from the PDK libraries. Since it is usually small we ignore it in our estimation in interest of runtime.

OpenSTA~\cite{opensta} by default uses On Chip Variation (OCV) mode~\cite{openstamanual} which adds additional pessimism to compensate for variations in real-world process, voltage and temperature. The tool also by default enables a feature called Clock Reconvergence Pessimism Removal (CRPR) in OCV mode~\cite{openstamanual}. When a part of clock path is common between the launch flip-flop and capture flip-flop, due to the additional pessimism in OCV, the STA for common clock path becomes overly pessimistic. This happens because the tool considers the maximum clock delay of the launch path for AT calculation, while it considers the minimum clock delay of capture path for RAT calculation. Both variations cannot be true simultaneously for the common part of the launch and capture path. To compensate for this extra pessimism, an extra CRP factor is added to our calculated RAT defined by:
$$
CRP = max(\text{common\_path\_delay}) - min(\text{common\_path\_delay})
$$
Since the model is trained on AT labels generated by OpenSTA with OCV and CRPR enabled, we need to consider it for RAT calculation too. CRPR is an expensive process, since it involves traversing the clock paths to determine common part. To minimize this cost, we do CRPR only for critical paths i.e. paths with negative slack. Since CRPR increases slack, it can only eliminate some critical paths by making their slack greater than zero. This makes sure that using uncorrected RAT (before CRPR) for identifying critical paths is valid, and it is impossible to miss any critical paths.

It is important to note that when presenting RAT/TNS/WNS results of our algorithm in Table \ref{tab:tnswns} and Table \ref{tab:results} we use AT labels (from post-routing STA) instead of predicted AT from TimingPredict. That is, we assume that the AT prediction model is ideal. This ensures that the results presented for RAT estimation are independent of the performance of the AT prediction model.

\section{Experiments}

\begin{table}[h]
  \resizebox{.45\textwidth}{!}{%
  \begin{tabular}{|c|cc|cc|cc|}
    \toprule
    & \multicolumn{2}{c|}{PostR STA} & \multicolumn{2}{c|}{Ours} && \\
    Circuit & TNS & WNS & TNS & WNS & $\Delta$ TNS & $\Delta$ WNS\\
    \midrule
    BM64 & -920.42 & -5.35 & -622.24 & -4.68 & 298.18& 0.67\\
    salsa20 & -26.96 & -2.46 & -22.78 & -2.32 & 4.18 & 0.14\\
    aes128 & -10943.44 & -10.86 & -9755.54 & -10.51 & 1187.9 & 0.35\\
    wbqspiflash & -18.81 & -3.26 & -13.24 & -2.95 & 5.57 & 0.31\\
    aes256 & -9599.43 & -9.78 & -7372.24 & -9.46 & 2227.19 & 0.32\\
    des & -1945.19 & -5.62 & -18.41 & -1.18 & 1926.78 & 4.44\\
    aes\_cipher & -1551.87 & -6.19 & -1153.56 & -5.81 & 398.31 & 0.38\\
    picorv32a & -565.20 & -20.16 & -521.18 & -19.95 & 44.02 & 0.21\\
    usb & -0.69 & -0.69 & -0.69	& -0.69	& 0 & 0\\
    PPU & -16.25 & -1.77 & -10.94 & -1.53 & 5.31 & 0.24\\
    chacha & -902.30 & -6.07 & -795.92 & -5.64 & 106.38 & 0.43\\
    jpeg\_encoder & -10571.76 & -14.42 & -8198.46 & -13.89 & 2373.3 & 0.53\\
    usbf\_device & -206.42 & -9.16 & -189.54 & -8.84 & 16.88 & 0.32\\
    aes192 & -14198.15 & -14.08 & -12394.23 & -13.93 & 1803.92 & 0.15\\
    xtea & -15.13 & -2.15 & -1.14 & -0.58 & 13.99 & 1.57\\
    y\_huff & -5.17 & -1.07 & 0.00 & 0.00 & 5.17 & 1.07\\
    point\_mult & -319.98 & -2.82 & -233.80 & -2.68 & 86.18 & 0.14\\
    y\_dct & -1806.23 & -5.71 & -1530.68 & -5.15 & 275.55& 0.56\\
  \bottomrule
\end{tabular}}
\vspace{1em}
\caption{Comparison of TNS/WNS calculated using our method vs calculated by OpenSTA~\cite{opensta} post-routing (ground truth). Only circuits with non-zero TNS/WNS are shown. For all remaining circuits with zero TNS/WNS our method also calculates zero.}
\label{tab:tnswns}
\end{table}

\subsection{Training Setup}
All experiments are performed on a Linux system with Ubuntu 22.04. The machine is equipped with a 24-core Intel (R) Core (TM) i7-13700KF processor and one 32GB GPU. 
\subsection{Dataset}
We use the 21 benchmark circuits provided by TimingPredict~\cite{guo2022timing} as well as 13 more circuits collected by us from examples made available in the OpenLane~\cite{openlane} repository. We run the OpenLane EDA tool on these examples to perform the synthesis-placement-routing flow and obtain the LIB, DEF and Verilog files. Not all of the circuits in the repository were usable since the flow failed for various reasons and were discarded. In case of TimingPredict dataset the authors already make these files available publicly. Next we use a modified version of OpenSTA~\cite{openstadump} to perform STA and obtain labels for AT, RAT, CD, ND and Slew. Finally we run our proposed TimingParser to convert all 34 circuits into circuit graphs. Note that although the TimingPredict dataset is also made available as pre-parsed graphs, we avoid using them to make sure that all the circuit graphs are consistent in format.

For TimingPredict dataset we use the same train/test split as used in the original work. For our OpenLane dataset, we randomly split it into 9 train circuits and 4 test circuits.

\subsection{Metrics}
For AT prediction, we use the R2 score to evaluate the results, which is the same metric used in previous work \cite{guo2022timing}. 
$$
R2 = 1 - \frac{MSE}{Variance}
$$
However, the R2 score is inappropriate for RAT since unlike AT, RAT has low variance across the endpoints of the circuit. This leads to misleadingly low R2 scores even when the actual error is low. Therefore, we present the Mean Absolute Error (MAE) for RAT in our results. The mean is across all endpoints for a given circuit.


\section{Results}
\subsection{RAT Prediction}

While several prior works have attempted to predict AT at pre-routing stage \cite{guo2022timing, zhong2024preroutgnn}, how to acquire pre-routing RAT is missing in these methods. Due to the lack of research for this task, we compare our RAT prediction module against two baselines: (1) TimingPredict model adapted for RAT prediction and (2) pre-routing STA tool.

The original TimingPredict model is only intended for AT prediction and it uses the following loss function:
$$
L = L_{CD} + L_{ND} + L_{AT|Slew}
$$
where all loss functions are Mean Squared Error (MSE) and '|' denotes concatenation. Cell Delay (CD) prediction and Net Delay (ND) prediction are auxiliary tasks.

We modify the model to also predict RAT along with AT and Slew so that the loss becomes:
$$
L = L_{CD} + L_{ND} + L_{AT|Slew|RAT}
$$
The original model is trained for 20000 epochs, while the modified model is trained for 100000 epochs accounting for the increased complexity of task. The checkpoint with best loss is used for evaluation on test dataset.

Pre-routing RAT can also be obtained from STA tool. Due to the lack of interconnect information prior to routing, the SPEF parasitic results are not available. To obtain pre-routing RAT, STA is performed without the \texttt{read\_spef} step shown in Figure \ref{fig:arch}.

The results of pre-routing RAT prediction are shown in Table \ref{tab:results}. Only the late-rise corner is considered for "RAT MAE", as that is what is used in setup slack calculation. "Ours" denote the RAT calculation using our proposed algorithm. "PreR STA" denotes the use of the OpenSTA~\cite{opensta} tool at the pre-routing stage. As demonstrated in the table, our proposed RAT prediction algorithm significantly outperforms the two baselines across the majority of circuits. The average MAE on test dataset is 0.6454, compared to 1.5165 for PreR STA and 8.1608 for modified TimingPredict, representing a reduction of at least \textbf{57.4\%}. 

\subsection{TNS/WNS Prediction}

With AT and RAT for each timing path, we can compute TNS/WNS metrics for each circuit, which are the two most important timing metrics used by engineers to evaluate circuit quality. In Table \ref{tab:tnswns} we show the TNS/WNS calculated by our method compared to the ground truth obtained by running the OpenSTA~\cite{opensta} tool post-routing. We use the STA commands \texttt{report\_tns} and \texttt{report\_wns} to obtain these groundtruth values. As shown in the table, our proposed E2ESlack tool can reliably predict WNS at pre-routing stage. TNS prediction is less accurate due to the accumulation of errors across all timing paths. Considering the significant runtime saving as detailed in the following section, our method can be used as a valuable early indicator for engineers regarding the circuit's timing performance.

\subsection{Runtime Analysis}

Finally, we analyze the runtime of using our framework, which includes parsing circuit data, predicting AT using an AT prediction model, and RAT calculation using proposed algorithm. This is a much more realistic analysis compared to prior work~\cite{guo2022timing}, which only considers model inference time for AT prediction and completely ignores the time taken for parsing and RAT estimation. In Table~\ref{tab:results}, the runtime of our framework is compared to the time consumption of routing and STA in the traditional EDA pipeline. On average, our method can save up to 23x in runtime to obtain TNS/WNS metrics, enabling faster design iteration.


\section{Conclusion}

Our work introduces an end-to-end graph-based framework for pre-routing slack prediction, addressing RAT estimation for the first time. Our framework also provides tools for extracting circuit features, constructing graphs and calculating TNS and WNS metrics. Extensive experiments demonstrate the accuracy and efficiency of our method outperforming both modified TimingPredict~\cite{guo2022timing} and pre-routing STA. The proposed framework can be potentially used as an early indicator of a circuit's timing performance, significantly accelerating circuit design process.


\bibliographystyle{ACM-Reference-Format}
\bibliography{reference}


\end{document}